\documentclass[10pt,twocolumn,letterpaper]{article}

\usepackage{cvpr}              

\usepackage{multirow}
\usepackage{multicol}
\usepackage{tabularx}
\usepackage{graphicx}
\usepackage{pifont}
\definecolor{darkgray}{gray}{0.3}

\definecolor{cvprblue}{rgb}{0.21,0.49,0.74}
\usepackage[pagebackref,breaklinks,colorlinks,allcolors=cvprblue]{hyperref}
\usepackage[accsupp]{axessibility}

\def\paperID{1563} 
\def\confName{CVPR}
\def\confYear{2025}

\title{STOP: Integrated Spatial-Temporal Dynamic Prompting for Video Understanding}

\author {
    Zichen Liu\textsuperscript{\rm 1},
    Kunlun Xu\textsuperscript{\rm 1},
    Bing Su\textsuperscript{\rm 2},
    Xu Zou\textsuperscript{\rm 3},
    Yuxin Peng\textsuperscript{\rm 1},
    Jiahuan Zhou\textsuperscript{\rm 1}\thanks{Corresponding author}    
    \\
    {\small \textsuperscript{\rm 1} Wangxuan Institute of Computer Technology, Peking University, Beijing, China}\\
    {\small \textsuperscript{\rm 2} Gaoling School of Artificial Intelligence, Renmin University of China, Beijing, China}\\
    {\small \textsuperscript{\rm 3} School of Artificial Intelligence and Automation, Huazhong University of Science and Technology, Wuhan, China}\\
    {\tt\small  \{lzc20180720, xkl\}@stu.pku.edu.cn, subingats@gmail.com, zoux@hust.edu.cn}\\  
    {\tt\small  \{pengyuxin, jiahuanzhou\}@pku.edu.cn}  
}

\begin{document}
\maketitle

\begin{abstract}
Pre-trained on tremendous image-text pairs, vision-language models like CLIP have demonstrated promising zero-shot generalization across numerous image-based tasks. However, extending these capabilities to video tasks remains challenging due to limited labeled video data and high training costs. Recent video prompting methods attempt to adapt CLIP for video tasks by introducing learnable prompts, but they typically rely on a single static prompt for all video sequences, overlooking the diverse temporal dynamics and spatial variations that exist across frames. This limitation significantly hinders the model’s ability to capture essential temporal information for effective video understanding. To address this, we propose an integrated \textbf{S}patial-\textbf{T}emp\textbf{O}ral dynamic \textbf{P}rompting (\textbf{STOP)} model which consists of two complementary modules, the intra-frame spatial prompting and inter-frame temporal prompting. Our intra-frame spatial prompts are designed to adaptively highlight discriminative regions within each frame by leveraging intra-frame attention and temporal variation, allowing the model to focus on areas with substantial temporal dynamics and capture fine-grained spatial details. Additionally, to highlight the varying importance of frames for video understanding, we further introduce inter-frame temporal prompts, dynamically inserting prompts between frames with high temporal variance as measured by frame similarity. This enables the model to prioritize key frames and enhances its capacity to understand temporal dependencies across sequences. Extensive experiments on various video benchmarks demonstrate that STOP consistently achieves superior performance against state-of-the-art methods. The code is available at \href{https://github.com/zhoujiahuan1991/CVPR2025-STOP}{https://github.com/zhoujiahuan1991/CVPR2025-STOP}.
\end{abstract}

\begin{figure}[htbp]
\begin{center}
\includegraphics[width=1\linewidth]{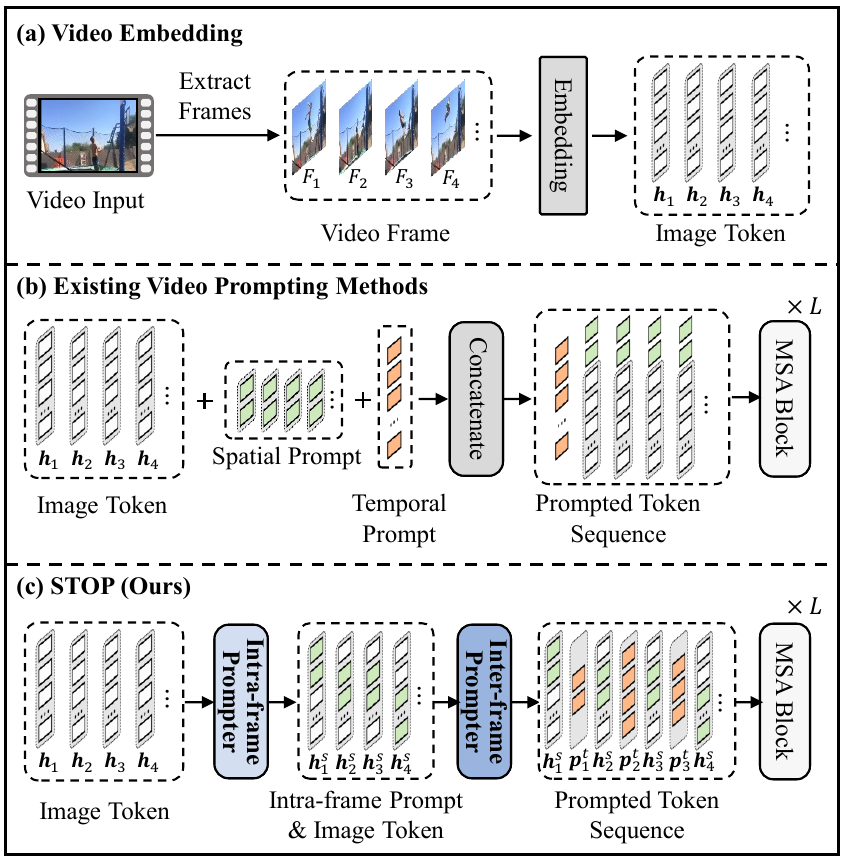}
\end{center}
\vspace{-10pt}
\caption{\label{fig:motivation} 
Existing video prompting methods~\cite{yang2024dgl, huang2023vop} typically add a static prompt across diverse videos, limiting the model's ability to capture essential temporal information. In contrast, our approach introduces dynamic intra-frame spatial prompts and inter-frame temporal prompts, guiding the model to focus on discriminative regions and key frames with significant temporal dynamics.
}
\end{figure}

\section{Introduction}
\label{sec:intro}
With the advancement of deep learning~\cite{radford2021learning, ye2022unsupervised, dai2023cross, huijuan2023improved, xu2024dask, xu2024mitigate, xu2024lstkc, xu2024distribution}, the data required for model training has been steadily increasing. Recently, vision-language models such as CLIP, which are pre-trained on large-scale image-text pairs by contrastive learning, have demonstrated remarkable zero-shot generalization capabilities in downstream tasks such as image classification and image-text retrieval~\cite{radford2021learning, zhou2022learning, zhou2022conditional}. However, applying this pre-training and zero-shot generalization paradigm to video tasks poses unique challenges. Compared to images, labeled video data is more limited, which makes it challenging to collect large-scale video-text pairs~\cite{luo2022clip4clip, zhang2024mpt}. Additionally, training large-scale video-language models through contrastive learning with video-text pairs entails high computational costs, making it challenging to apply training paradigms similar to CLIP for video-language tasks~\cite{yang2024dgl, huang2023vop}.

To address this issue, recent studies~\cite{huang2023vop, yang2024dgl, zhang2024mpt} have focused on efficiently fine-tuning large-scale pre-trained vision-language models on video data, enabling them to handle downstream video understanding tasks at a lower cost. Among these methods, video prompting, or video prompt learning, has emerged as an important category of approaches. It adapts vision-language models to video tasks by introducing learnable video prompts while keeping the model's pre-trained parameters fixed. However, as shown in Figure.~\ref{fig:motivation}, existing video prompting methods~\cite{huang2023vop, yang2024dgl, zhang2024mpt} learn a single static prompt for all data, overlooking the differences in keyframes with significant temporal dynamics and the varying discriminative regions within each frame. This results in inaccuracies in the video frames and regions that the pre-trained vision-language model focuses on. As a consequence, the model's ability to capture temporal information is limited, which ultimately hinders its capacity to effectively understand video content.

To address this issue, we propose an integrated \textbf{S}patial-\textbf{T}emp\textbf{O}ral dynamic \textbf{P}rompting approach, named \textbf{STOP}. Regions with temporal dynamics are essential for understanding video actions, yet image-text pairs pre-trained CLIP models often struggle to attend effectively to these areas. To overcome this limitation, we introduce the Intra-frame Spatial Prompting. Our approach incorporates a lightweight 3D convolutional network to capture the temporal dynamics of different regions of the video. By combining this information with intra-frame attention weights, we identify discriminative regions within videos that include both primary objects in single frames and dynamic temporal information. Building on these insights, we design an intra-prompt prompter to generate spatial prompts for these regions, guiding the model to focus on areas with significant temporal changes and thereby enhancing its ability to capture fine-grained critical information in video data.

Furthermore, Considering that the dynamic changes vary across frames, affecting their importance for video understanding, we propose inter-frame temporal prompting to help the pre-trained model focus on keyframes. Building on the intra-frame spatial prompting that identifies discriminative regions within video frames, we further calculate the degree of change in these regions between frames. For keyframes with significant temporal dynamics, we dynamically generate inter-frame prompts using a lightweight prompter and insert them to provide fine-grained information between two frames. The intra-frame spatial prompts and inter-frame temporal prompts complement each other, guiding the model to focus on critical spatial and temporal locations, thereby enhancing the model's ability to accurately understand videos, leading to improved performance.

To sum up, the main contributions of this work are:
(1) To address the challenge of pre-trained vision-language models struggling to capture temporal information in videos, we propose the STOP method. First, we design intra-frame spatial prompting to highlight discriminative regions in video frames, effectively guiding the model to focus on dynamically changing areas.
(2) Furthermore, we calculate the dynamic changes between video frames in the video and dynamically generate inter-frame temporal prompts. These prompts are inserted between frames with significant dynamic changes, providing fine-grained temporal information to facilitate the model focusing on and understanding the keyframes in the video.
(3) Extensive experiments on various benchmarks for video action recognition and video-text retrieval demonstrate that our proposed STOP method consistently and significantly outperforms existing video prompting methods.


\section{Related Work}
\label{sec:related}
\subsection{Vision-Language Pre-training}
Vision-language pre-training (VLP) aims to learn joint representations of vision and language, achieving strong performance across various downstream tasks~\cite{bin2024gallerygpt, mitra2024compositional, chang2024survey, radford2021learning}. Recent advancements in contrastive image-text pre-training, such as CLIP~\cite{radford2021learning}, leveraging large-scale image-text pairs from the internet for training. Works like Flamingo~\cite{alayrac2022flamingo} and ALBEF~\cite{li2021align} further demonstrate the effectiveness of pre-trained models. For video-language pre-training, models like CLIPBERT~\cite{lei2021less} and Frozen in Time~\cite{bain2021frozen} have shown promise, yet challenges persist, including limited video datasets and high computational costs. To address these, methods like CLIP4Clip~\cite{luo2022clip4clip} and X-Pool~\cite{gabeur2020multi} transfer image-text pre-trained models to video tasks, though they still require fine-tuning all model parameters, resulting in high overhead.

\begin{figure*}[ht]
\begin{center}
\includegraphics[width=1\linewidth]{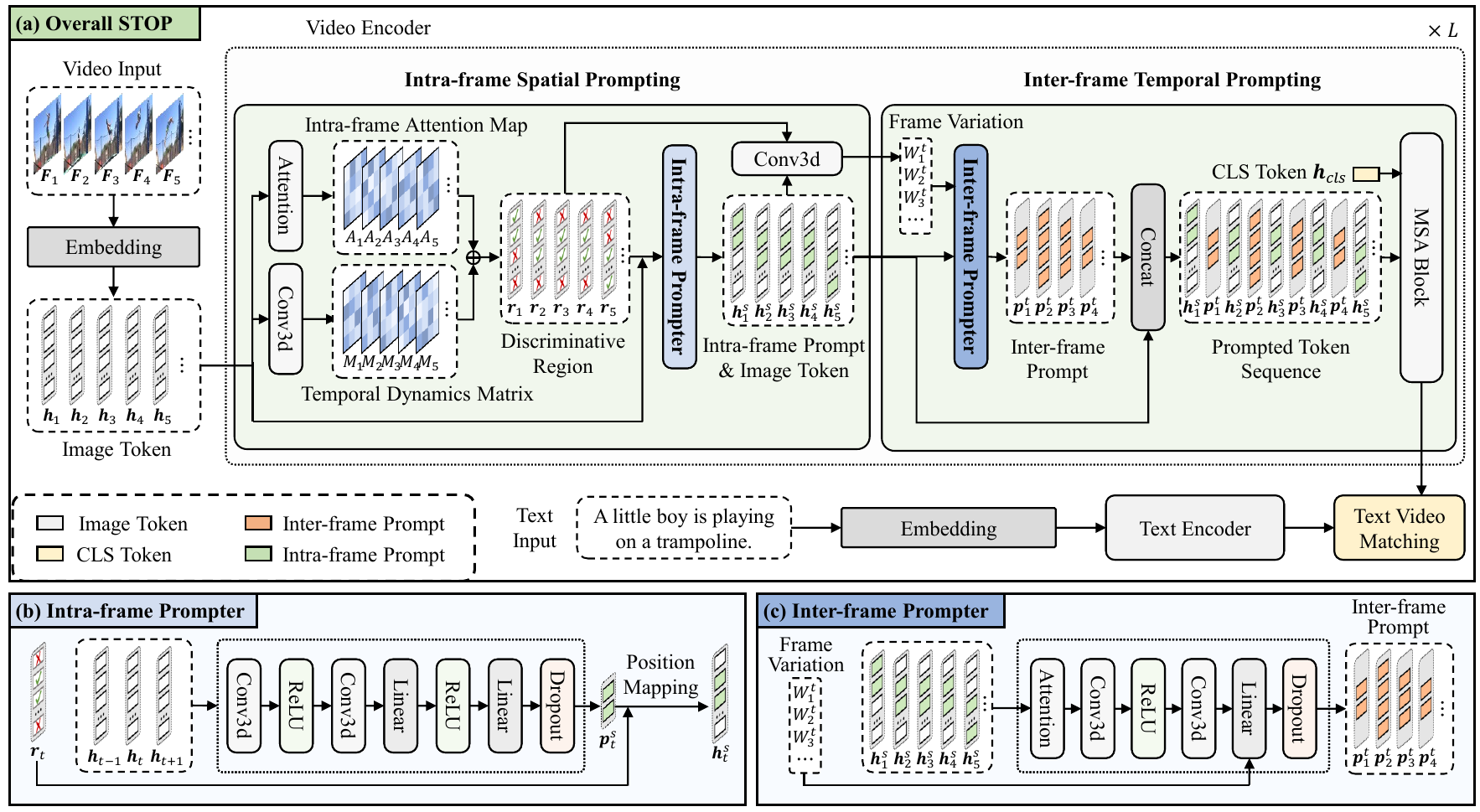}
\end{center}
\vspace{-10pt}
\caption{\label{fig:framework} 
    The pipeline of our STOP. For each video, we begin by embedding it into image tokens. Then the intra-frame spatial prompting is introduced to locate discriminative regions and add generated prompts to these areas. Based on the degree of inter-frame variation, we dynamically generate inter-frame prompts and insert them as needed. Finally, these prompts along with the CLS token, pass through the MSA (multi-head self-attention) block to obtain a video representation, which is then used to compute similarity with the text features.
}
\end{figure*}

\subsection{Video Prompting}

Prompt learning, originating in NLP, aims to adapt pre-trained language models to various tasks~\cite{lester2021power, li2021prefix, liu2021p}, by introducing learnable tokens as prompts, optimizing only these tokens during training. Inspired by prompt learning's success in NLP, this approach extends to vision-language models (VLMs)~\cite{zhou2022learning, zhou2022conditional, khattak2023maple, khattak2023self, liu2024dart} and visual models~\cite{jia2022visual, liu2024insvp, liu2024compositional,li2024exemplar}. For instance, CoOp~\cite{zhou2022learning} uses learnable text prompts for improved classification performance, while VPT~\cite{jia2022visual} introduces token-level prompts in the vision branch to capture image attributes. 

 Recently, prompt learning has been expanded to video tasks, like video understanding and text-video retrieval~\cite{huang2023vop, yang2024dgl, diao2024unipt, zhang2024mpt}. For instance, VoP~\cite{huang2023vop} innovates with visual prompts to capture spatiotemporal video characteristics, and DGL~\cite{yang2024dgl} enhances video-text interaction through global-local prompt coordination. Similarly, MPT~\cite{zhang2024mpt} refines prompts to mine more detailed modality-specific information, boosting retrieval performance. However, these methods apply static prompts to all videos, neglecting video-specific frame details and dynamic changes, which limits the model’s ability to capture temporal nuances and constrains overall performance.

\subsection{Video Understanding}
Video action recognition and video-text retrieval are two main research areas related to video understanding. For video action recognition, researchers have focused on advancing temporal modeling efficiency. Optical flow methods, often used in two-stream fusion, are effective but computationally demanding~\cite{simonyan2014two, wang2016temporal}. An alternative approach, 3D convolutions, extends 2D convolutions into the spatiotemporal domain but also suffers from high complexity~\cite{tran2015learning, carreira2017quo, tran2018closer}. Other strategies embed plug-and-play temporal modules into 2D networks or adapt LSTMs for sequence analysis~\cite{lin2019tsm, liu2021tam}. Recently, Vision Transformer-based architectures have shown promising results in temporal modeling and feature extraction efficiency~\cite{bertasius2021space, liu2022video, li2022uniformer}. In addition to single-modality methods, multi-modal techniques leverage CLIP's capabilities for tasks like video classification and retrieval~\cite{wang2021actionclip, luo2022clip4clip, ju2022prompting}.

For video-text retrieval task, early multi-modal researches~\cite{liu2019use, gabeur2020multi} focused on feature fusion for retrieval enhancement, while more recent studies~\cite{chen2020fine, wu2021hanet} emphasize cross-modal alignment at levels such as events, actions, and entities. Inspired by large-scale pre-trained models like BERT~\cite{kenton2019bert} and CLIP~\cite{radford2021learning}, contemporary methods~\cite{liu2022contrastive} increasingly employ contrastive learning or masked language modeling to develop a shared video-text representation space. In our work, we adapt the frozen CLIP model with learnable prompts, effectively transferring its strengths to video retrieval, as validated by our empirical results.

\section{Method}
\label{sec:method}
In this section, we illustrate the proposed \textit{STOP} in detail, and the overall pipeline is depicted in Figure~\ref{fig:framework}.
\subsection{Notations}
The backbone of our method is CLIP~\cite{radford2021learning}, with CLIP4clip~\cite{luo2022clip4clip} as the baseline. It mainly consists of two branches: the text encoder $\boldsymbol{\mathrm{E}}_T$ and the video encoder $\boldsymbol{\mathrm{E}}_V$. The pre-trained text encoder $\boldsymbol{\mathrm{E}}_T$ is a transformer model~\cite{vaswani2017attention}. For the video-text retrieval task, the input of the text encoder $\boldsymbol{\mathrm{E}}_T$ is a natural language sentence $\boldsymbol{S} $. For the action recognition task, given a video’s category text description $C = \{c_1, c_2, \dots, c_{K}\}$, where $K$ denotes the number of categories, we construct a sentence using a hand-crafted prompt following \cite{wang2024vilt}, such as $\boldsymbol{S}=\text{“A video of the action [CLS]”}$. Then, we use the text encoder $\boldsymbol{\mathrm{E}}_T$ to obtain the representation $\boldsymbol{s}=\boldsymbol{\mathrm{E}}_T(\boldsymbol{S})$ of the sentence $\boldsymbol{S}$, where $\boldsymbol{s} \in \mathbb{R}^{d}$ and $d$ is the dimension of representation.

The image encoder in CLIP and CLIP4clip is based on the Vision Transformer (ViT)~\cite{dosovitskiy2020image}. The input is a video $\boldsymbol{V} \in \mathbb{R}^{N_F \times 3 \times H \times W}$, where $N_F$ represents the number of frames and \( H \times W \) is the spatial size. Each video frame $\{ \boldsymbol{F}_i \}_{i=1}^{N_F}$ is split into \( N_p = \frac{H \times W}{h\times w} \) fixed-size patches of size \( h \times w \), and these patches are flattened into a set of vectors \( \boldsymbol{x}_i = \{ \boldsymbol{x}_{i,j} \in \mathbb{R}^{h \times w} \}_{j=1}^{N_p} \), where \( i \) denotes the frame index and \( j \) denotes the patch index. These vectors are then projected into patch embeddings $ \boldsymbol{h}_i = \{ \boldsymbol{h}_{i,j} \}_{j=1}^{N_p} $, where $\boldsymbol{h}_{i,j} \in \mathbb{R}^{d_v}$ and $d_v$ is the embedding dimension. Then all embeddings $ \{ \boldsymbol{h}_i \}_{i=1}^{N_F}$ and the CLS token $\boldsymbol{h}_{cls} \in \mathbb{R}^{d_v}$ are then input to the video encoder $\boldsymbol{\mathrm{E}}_V$, getting the representation $\boldsymbol{v} \in \mathbb{R}^{d}$. The video representation $\boldsymbol{v}$ and text representation $\boldsymbol{s}$ are then used to calculate cosine similarity, producing the retrieval or action recognition results.

\subsection{Integrated Spatial-Temporal Dynamic Prompting}
\label{subsec:STOP}
Our integrated spatial-temporal dynamic prompting method includes two modules, intra-frame spatial prompting and inter-frame temporal prompting. They are functionally complementary,  guiding the pre-trained vision-language model to accurately focus on discriminative regions of the video in the spatial and temporal dimensions, respectively. The intra-frame spatial prompting module first identifies the location of the discriminative region $\boldsymbol{r}_i \in \mathbb{R}^{N_p} $ and then generates intra-frame spatial prompts $\boldsymbol{p}^s_i$ for frame $\boldsymbol{F}_i$ through a lightweight prompter $\mathcal{P}^s$:
\begin{equation}
    \boldsymbol{p}^s_i = \mathcal{P}^s(\boldsymbol{h}_{i-1}, \boldsymbol{h}_i, \boldsymbol{h}_{i+1}) ,
    \label{eq:spatial-prompter}
\end{equation}
Then, the intra-frame spatial prompt $\boldsymbol{p}^s_i$ is overlaid onto the discriminative region $\boldsymbol{r}_i$ of $\boldsymbol{h}_{i}$, resulting in $\boldsymbol{h}_{i}^{s}$.

The inter-frame video prompting module first utilizes the discriminative regions obtained by the intra-frame video prompting module and a 3D convolutional network to compute the temporal variation $\{w_i\}_{i=1}^{N_F-1}$ between adjacent frames in the video. $w_t \in \mathbb{R}$ represents the degree of temporal variation between frames $\boldsymbol{F}_i$ and $\boldsymbol{F}_{i+1}$. Then, we use an inter-frame prompter $\mathcal{P}^t$ to generate inter-frame prompts $\{\boldsymbol{p}^t_i\}_{i=1}^{N_F-1}$ with $\{\boldsymbol{h}_{i}^{s}\}_{i=1}^{N_F}$ and $\{\boldsymbol{r}_{i}\}_{i=1}^{N_F}$. Then, we concatenate the inter-frame prompts $\{\boldsymbol{p}^t_i\}_{i=1}^{N_F-1}$ with the image token $\{\boldsymbol{h}_{i}^{s}\}_{i=1}^{N_F}$ and use them as the input to the pre-trained model's MSA block.

\subsection{Intra-frame Spatial Prompting}
As mentioned in Sec~\ref{subsec:STOP}, we first identify the discriminative region $\boldsymbol{r}_i$ of each frame $\boldsymbol{F}_i$. Specifically, we compute the attention map $A_i$ of different patches $\boldsymbol{h}_i = \{ \boldsymbol{h}_{i,j} \in \mathbb{R}^{d_v} \}_{j=1}^{N_p}$ of a single frame $\boldsymbol{F}_i$ using the self-attention module $\mathrm{Attn}(\cdot)$ of the pre-trained model:
\begin{equation}
    A_i = \mathrm{Attn}(\boldsymbol{h}_{cls}, \boldsymbol{h}_i),
\end{equation}
where $A_i \in \mathbb{R}^{N_p}$. Additionally, a 3D convolutional layer $\mathcal{N}^s$ is utilized to calculate the temporal dynamics along the temporal dimension:
\begin{equation}
    [\tilde{\boldsymbol{h}}_1, \tilde{\boldsymbol{h}}_2, \cdots, \tilde{\boldsymbol{h}}_{N_F}] = \mathcal{N}^s([\boldsymbol{h}_1, \boldsymbol{h}_2, \cdots, \boldsymbol{h}_{N_F}] )
\end{equation}
\begin{equation}
    M_{i, j} = \frac{1}{d_v} \sum_{k}{\tilde{\boldsymbol{h}}_{i,j,k}^2},
\end{equation}
where $\tilde{\boldsymbol{h}_i} \in \mathbb{R}^{N_p \times d_v}$ and $M_{i, j} \in \mathbb{R}$ represents the temporal dynamic of the $j$-th token of the $i$-th frame. $[\;]$ denotes the concatenation operation.

Then, we compute the discriminative regions $\boldsymbol{r}_i$ of each frame by using the attention map $A_i$ and the temporal dynamic variation $M_i$:
\begin{equation}
    W_i^s = \alpha A_i + (1-\alpha) M_i,
\end{equation}
where $\alpha$ is a weight hyper-parameter. Then, we get the discriminative regions of each video frame with \( N_s \) patches:
\begin{equation}
    \boldsymbol{r}_{i,j} = 
    \begin{cases} 
        1,  & \text{if }W_{i,j}^s\text{ is the top }N_s \text{ largest values of }W_i^s\\
        0, & \text{otherwise }
    \end{cases}
    ,
\end{equation}
where $N_s$ is a hyper-parameter.

Through the above process, we comprehensively consider the degree of temporal variation and the importance of understanding individual frames, thereby obtaining the discriminative regions of the video. Next, as introduced in Eq.~\ref{eq:spatial-prompter} of Sec.~\ref{subsec:STOP}, we use a lightweight prompter $\mathcal{P}^s$ to generate intra-frame spatial prompts $\boldsymbol{p}^s_i$, which we then overlay onto the tokens corresponding to the discriminative regions:
\begin{equation}
    \boldsymbol{h}_{i,j}^{s} = \boldsymbol{h}_{i,j} + \boldsymbol{r}_{i,j} \cdot \boldsymbol{p}^s_{i,j} .
\end{equation}
With the intra-frame spatial prompting, we identify the discriminative regions of each video frame and add prompts to the corresponding tokens, guiding the pre-trained model to focus accurately.

\subsection{Inter-frame Temporal Prompting}
Building on the intra-frame spatial prompting that highlights the discriminative regions within frames, we further introduce inter-frame temporal prompting to identify key frames along the temporal dimension. As mentioned in Sec.~\ref{subsec:STOP}, we first use a 3D convolutional layer $\mathcal{N}^t$ to obtain the dynamic variation between adjacent frames:
\begin{equation}
    \Delta\boldsymbol{h}_i^s = \boldsymbol{h}_i^s - \boldsymbol{h}_{i-1}^s,
\end{equation}
\begin{equation}
    [\tilde{\boldsymbol{h}}_1^{s}, \cdots, \tilde{\boldsymbol{h}}_{N_F-1}^{s}] = \mathcal{N}^t([\Delta\boldsymbol{h}_1^s, \cdots, \Delta\boldsymbol{h}_{N_F-1}^s)]) ,
\end{equation}
\begin{equation}
    W^t_{i} = \frac{1}{N_p \cdot d_v} \sum_{j}{( (1+\beta \cdot \boldsymbol{r}_{i,j}) \sum_{k}{(\tilde{\boldsymbol{h}}_{i,j,k}^s)^2} )} ,
\end{equation}
where $\tilde{\boldsymbol{h}}^s_i \in \mathbb{R}^{N_p \times d_v}$, $W_i^t \in \mathbb{R}$ and $\beta$ is a weight hyper-parameter.
Through the above process, when calculating frame variation, we assign higher weights to the discriminative regions $\boldsymbol{r}_i$ identified by intra-frame spatial prompting, enabling a more accurate assessment of changes of the main object rather than the background.

Then, based on the magnitude of the frame temporal variation $W_i^t$, we determine the number of prompt tokens $N^t_i \in \mathbb{R}$ to be inserted between frames:
\begin{equation}
    N^t_i = \lceil \eta \cdot W_i^t \rceil ,
\end{equation}
where $\lceil \cdot \rceil$ denotes the ceiling (rounding up to the nearest integer) and $\eta$ is a scaling factor. Next, we introduce an inter-frame prompter $\mathcal{P}^t$ to generate the inter-frame prompts $\boldsymbol{p}^t_i$:
\begin{equation}
    [\boldsymbol{p}_1^t, \cdots, \boldsymbol{p}_{N_F}^t] = \mathcal{P}^t([\Delta\boldsymbol{h}_1^s, \cdots, \Delta\boldsymbol{h}_{N_F-1}^s]) ,
\end{equation}
where $\boldsymbol{p}_i^t \in \mathbb{R}^{N^t_i \times d_v}$. For different prompt token numbers $N_i^t$, the final linear layer of the prompter corresponds to selecting a net with an output dimension of $N^t_i$. Then, we concatenate the inter-frame temporal prompts $\boldsymbol{p}^t_i$ with the image tokens $\tilde{\boldsymbol{h}}_i^t$ and the CLS token $\boldsymbol{p}_{cls}$, and input them into the multi-head self-attention (MSA) block of the pre-trained model to obtain the final video representation $\boldsymbol{v} \in \mathbb{R}^d$. Then, we calculate the semantic relevance between $\boldsymbol{v}$ and the text representation $\boldsymbol{s}$ using cosine similarity:
\begin{equation}
    c(\boldsymbol{s}, \boldsymbol{v}) = \frac{\boldsymbol{s} \cdot \boldsymbol{v}}{\|\boldsymbol{s}\| \| \boldsymbol{v}\| },
\end{equation}
where $\|\cdot\|$ denotes the $\ell_2$ norm.

\subsection{Overall Optimization}

For the video action recognition task, following ~\cite{zhou2022learning, zhou2022conditional, wang2024vilt}, we use the cross-entropy loss function for training:
\begin{equation}
    \mathcal{L}_{act} = - \frac{1}{B} \sum_{i=1}^B{\log \frac{e^{c(\boldsymbol{v}_i, \boldsymbol{s}_{y_i}) / \tau}}{\sum_{j=1}^{K} e^{c(\boldsymbol{v}_i, \boldsymbol{s}_j) / \tau}}},
\end{equation}
where $y_i$ is the label of video $\boldsymbol{v}_i$, $B$ denotes the batch size and \( \tau \) is a temperature parameter. 

For the video-text retrieval task, following ~\cite{yang2024dgl, zhang2024mpt}, we use contrastive loss for training. We treat paired text-video data as positive samples, while other data in the batch are negative samples:
\begin{equation}
    \mathcal{L}_{vt} = \frac{1}{2B} \sum_{i=1}^{B} ( \log{\frac{e^{c(\boldsymbol{s}_{i}, \boldsymbol{v}_{i}) / \tau}}{\sum_{j} e^{c(\boldsymbol{s}_{j}, \boldsymbol{v}_{i}) / \tau}}} + \log{\frac{e^{c(\boldsymbol{s}_{i}, \boldsymbol{v}_{i}) / \tau}}{\sum_{j} e^{c(\boldsymbol{s}_{i}, \boldsymbol{v}_{j}) / \tau}}} ),
\end{equation}
In our proposed STOP, the parameters to be optimized include two 3D convolutional layers $\mathcal{N}^s$ and $\mathcal{N}^t$, and two promoters $\mathcal{P}^s$ and $\mathcal{P}^t$.

\begin{table}[b]
\small
  \centering
  \setlength{\tabcolsep}{2.5pt} 
  \begin{tabularx}{0.48\textwidth}{c|>{\raggedright\arraybackslash}m{3.2cm} | >{\centering\arraybackslash}m{1.5cm}>{\centering\arraybackslash}m{1.3cm}>{\centering\arraybackslash}m{1.2cm}}
    \toprule
    & Methods & HMDB51 & UCF101 & SS-V2\\
    \midrule
    & CLIP4Clip~\cite{luo2022clip4clip}                    
    & 75.2 & 94.1 & 69.4 \\
    \midrule
    \multirow{5}{*}{\rotatebox{90}{Adapter}} 
    & Bias~\cite{cai2020tinytl} 
    & 60.1 & 85.6 & 13.6\\
    & Adapter\textsuperscript{ATTN}~\cite{he2022towards} 
    & 60.3 & 85.7 & 14.2 \\
    & Adapter\textsuperscript{FFN}~\cite{chen2022adaptformer} 
    & 60.6 & 86.2 & 13.8 \\
    & Visual-Text Adapter~\cite{houlsby2019parameter} 
    & 63.4 & 88.3 & 14.6\\
    & Video-Text Adapter~\cite{pan2022st} 
    & 64.7 & 89.2 & 15.7\\
    \midrule  
    \multirow{5}{*}{\rotatebox{90}{Prompt}}       
    & VoP\textsuperscript{F+C}~\cite{huang2023vop} 
    & 65.2 & 91.3 & 16.7\\          
    & DGL-Linear~\cite{yang2024dgl} 
    & 67.2 & 92.5 & \textcolor{blue}{\textbf{18.3}}\\
    & DGL-Transformer~\cite{yang2024dgl}         
    & \textcolor{blue}{\textbf{69.8}} & \textcolor{blue}{\textbf{93.6}} & 18.1\\
    & UniPT~\cite{diao2024unipt}
    & 65.2 & 90.6 & 15.6\\
    & \textbf{STOP (Ours)} 
    & \textcolor{red}{\textbf{72.0}} & \textcolor{red}{\textbf{95.3}} & \textcolor{red}{\textbf{21.4}}\\
    \bottomrule
  \end{tabularx}
  \vspace{-5pt}
  \begin{flushleft}
        \footnotesize{- The best and second best results are marked in \textcolor{red}{\textbf{RED}} and \textcolor{blue}{\textbf{BLUE}}, respectively. }
    \end{flushleft}
    \vspace{-10pt}
  \caption{Comparison with state-of-the-art on the HMDB51, UCF101 and SS-V2. Here, we report the action recognition classification accuracy (ACC@1). For a fair comparison, all methods use CLIP-ViT-B/32~\cite{radford2021learning} as the backbone.}
  \label{tab:action}
  \vspace{-5pt}
\end{table}

\begin{table*}
\small
  \centering
  \setlength{\tabcolsep}{2.5pt} 
  \begin{tabularx}{\textwidth}{c|>{\raggedright\arraybackslash}m{3.2cm}>{\raggedright\arraybackslash}m{2.1cm}|>{\centering\arraybackslash}m{1cm} | >{\centering\arraybackslash}X>{\centering\arraybackslash}X>{\centering\arraybackslash}X>{\centering\arraybackslash}X | >{\centering\arraybackslash}X>{\centering\arraybackslash}X>{\centering\arraybackslash}X>{\centering\arraybackslash}X }
    \toprule
    & \multirow{2}{*}{Methods} & \multirow{2}{*}{} & Params & \multicolumn{4}{c|}{Text $\rightarrow$ Video} & \multicolumn{4}{c}{Video $\rightarrow$ Text}\\
    & & & (MB) $\downarrow$ & R@1 $\uparrow$ & R@5 $\uparrow$ & R@10 $\uparrow$ & MnR $\downarrow$ & R@1 $\uparrow$ & R@5 $\uparrow$ & R@10 $\uparrow$ & MnR $\downarrow$\\
    \midrule
    & CLIP4Clip~\cite{luo2022clip4clip} & \scriptsize{\textcolor{darkgray}{\textit{\textbf{Neurocomputing'22}}}}                   
    & 123.54& 43.1  & 70.4  & 80.8  & 16.2  & 43.1  & 70.5  & 81.2  & 12.4 \\
    \midrule
    \multirow{5}{*}{\rotatebox{90}{Adapter}} 
    & Bias~\cite{cai2020tinytl} & \scriptsize{\textcolor{darkgray}{\textit{\textbf{NeurIPS'20}}}}                   
    & 0.1   & 39.7  & 66.5  & 77.3  & 17.3  & 41.1  & 68.4  & 79.2  & 13.6 \\
    & Adapter\textsuperscript{ATTN}~\cite{he2022towards} & \scriptsize{\textcolor{darkgray}{\textit{\textbf{ICLR'22}}}}
    & 2.0   & 37.6  & 63.2  & 75.8  & 18.7  & 39.6  & 66.5  & 76.8  & 14.7 \\
    & Adapter\textsuperscript{FFN}~\cite{chen2022adaptformer} & \scriptsize{\textcolor{darkgray}{\textit{\textbf{NeurIPS'22}}}}
    & 2.0   & 38.2  & 63.5  & 76.4  & 17.9  & 39.9  & 66.8  & 77.7  & 14.2 \\
    & Visual-Text Adapter~\cite{houlsby2019parameter} & \scriptsize{\textcolor{darkgray}{\textit{\textbf{ICML'19}}}}
    & 11.82 & 39.2  & 65.7  & 76.1  & 17.6  & 40.7  & 68.8  & 77.6  & 13.7 \\
    & Video-Text Adapter~\cite{pan2022st} & \scriptsize{\textcolor{darkgray}{\textit{\textbf{NeurIPS'22}}}}
    & 11.94 & 41.1  & 67.0  & 77.1  & 17.4  & 42.6  & 68.4  & 78.4  & 13.8 \\
    \midrule  
    \multirow{10}{*}{\rotatebox{90}{Prompt}} 
    & Efficient Prompt~\cite{ju2022prompting} & \scriptsize{\textcolor{darkgray}{\textit{\textbf{ECCV'22}}}}                 
    & 6.35  & 36.7  & 64.6  & -     & -     & -     & -     & -     & -    \\          
    & VPT~\cite{jia2022visual} & \scriptsize{\textcolor{darkgray}{\textit{\textbf{ECCV'22}}}}                 
    & 0.18  & 42.0  & 66.6  & 77.3  & 19.2  & 39.4  & 66.8  & 77.2  & 16.2 \\
    & UPT~\cite{zang2022unified} & \scriptsize{\textcolor{darkgray}{\textit{\textbf{arXiv'22}}}}
    & 9.57  & 42.1  & 67.7  & 78.2  & 16.5  & 42.6  & 70.3  & 79.3  & 12.3 \\
    & VoP\textsuperscript{F+C}~\cite{huang2023vop} & \scriptsize{\textcolor{darkgray}{\textit{\textbf{CVPR'23}}}}  
    & 14.10 & 44.6  & 69.9  & 80.3  & 16.3  & 44.5  & 70.7  & 80.6  & \textcolor{red}{\textbf{11.5}} \\          
    & DGL-Linear~\cite{yang2024dgl} & \scriptsize{\textcolor{darkgray}{\textit{\textbf{AAAI'24}}}}                 
    & 0.83  & 44.7  & 70.5  & 79.2  & 16.2  & 42.1  & 70.0  & 80.6  & 13.4 \\
    & DGL-Transformer~\cite{yang2024dgl} & \scriptsize{\textcolor{darkgray}{\textit{\textbf{AAAI'24}}}}                 
    & 9.57  & 45.8  & 69.3  & 79.4  & 16.3  & 43.5  & 70.5  & 80.7  & 13.1 \\
    & UniPT~\cite{diao2024unipt} & \scriptsize{\textcolor{darkgray}{\textit{\textbf{CVPR'24}}}}                 
    & 9.60  & 38.9  & 60.2  & 71.4  & 18.5  & 39.3  & 58.6  & 70.4  & 16.4 \\
    & MPT-Linear~\cite{zhang2024mpt} & \scriptsize{\textcolor{darkgray}{\textit{\textbf{ACMMM'24}}}}                
    & 0.87  & 45.0  & 70.8  & 79.6  & 16.2  & 42.8  & 70.6  & \textcolor{red}{\textbf{81.1}} & 12.9 \\
    & MPT-Transformer~\cite{zhang2024mpt} & \scriptsize{\textcolor{darkgray}{\textit{\textbf{ACMMM'24}}}}                
    & 9.61  & \textcolor{blue}{\textbf{46.3}}  & \textcolor{blue}{\textbf{70.9}}  & \textcolor{blue}{\textbf{80.7}}  & \textcolor{blue}{\textbf{15.6}}  & \textcolor{blue}{\textbf{45.0}}  & \textcolor{blue}{\textbf{70.9}}  & 80.6  & 12.7 \\
    & \textbf{STOP (Ours)} & \scriptsize{\textcolor{darkgray}{\textit{\textbf{This Paper}}}}                
    & 7.53  & \textcolor{red}{\textbf{47.7}}  & \textcolor{red}{\textbf{71.4}}  & \textcolor{red}{\textbf{81.1}}  & \textcolor{red}{\textbf{15.2}}  & \textcolor{red}{\textbf{46.1}}  & \textcolor{red}{\textbf{71.6}}  & \textcolor{blue}{\textbf{81.0}}  & \textcolor{blue}{\textbf{12.2}} \\
    \bottomrule
  \end{tabularx}
  \begin{flushleft}
        \footnotesize{- ``Params'' represents the number of trainable parameters of each method. }\\
        \footnotesize{- The best and second best results are marked in \textcolor{red}{\textbf{RED}} and \textcolor{blue}{\textbf{BLUE}}, respectively. }
    \end{flushleft}
  \caption{Comparison with state-of-the-art on the MSR-VTT dataset. 
  For a fair comparison, all methods use CLIP-ViT-B/32~\cite{radford2021learning} as the backbone.
  }
  \label{tab:msrvtt}
\end{table*}

\section{Experiment}
\label{sec:experiment}
\subsection{Datasets}
\hspace*{1em}
\textbf{Action Recognition.} 
HMDB51~\cite{kuehne2011hmdb} includes around 7,000 clips across 51 action classes, featuring diverse real-world actions with variations in background and camera angles. UCF101~\cite{soomro2012ucf101} offers over 13,000 clips spanning 101 classes of activities, covering sports, daily actions, and social interactions, adding scene diversity. SS-V2 (Something-Something V2)~\cite{goyal2017something} has approximately 220,000 clips in 174 classes, emphasizing temporally dependent actions suitable for evaluating temporal reasoning.

\textbf{Text-Video Retrieval. }
We evaluate on four widely used text-video retrieval datasets: MSR-VTT~\cite{xu2016msr}, ActivityNet~\cite{caba2015activitynet}, DiDeMo~\cite{anne2017localizing}, and VATEX~\cite{wang2019vatex}. MSR-VTT contains 10,000 clips with around 20 descriptions each. ActivityNet includes 20,000 YouTube videos across 200 activities, with video descriptions combined into a single paragraph for video-text retrieval. DiDeMo consists of 10,000 Flickr videos paired with 40,000 sentences, using combined descriptions for model evaluation. The VATEX dataset is a large-scale video-text dataset containing 41,250 videos with bilingual English-Chinese captions, widely used for video captioning and cross-modal learning.

\textbf{Evaluation Metrics. } For video action recognition, we use classification accuracy (ACC@1) as the evaluation metric. For video-text retrieval, following previous work ~\cite{yang2024dgl, zhang2024mpt}, we use standard retrieval metrics to assess model performance, including R@K (Recall at Rank K, higher is better ↑) and MnR (Mean Rank, lower is better ↓).

\begin{table*}
\small
  \centering
  \setlength{\tabcolsep}{2.5pt} 
  \begin{tabularx}{\textwidth}{c|>{\raggedright\arraybackslash}m{3.2cm} | >{\centering\arraybackslash}m{0.9cm}>{\centering\arraybackslash}m{0.9cm}>{\centering\arraybackslash}m{1.1cm}>{\centering\arraybackslash}m{0.9cm} | >{\centering\arraybackslash}m{0.9cm}>{\centering\arraybackslash}m{0.9cm}>{\centering\arraybackslash}m{1.1cm}>{\centering\arraybackslash}m{0.9cm} | >{\centering\arraybackslash}m{0.9cm}>{\centering\arraybackslash}m{0.9cm}>{\centering\arraybackslash}m{1.1cm}>{\centering\arraybackslash}m{0.9cm} }
    \toprule
    & \multirow{2}{*}{Methods} & \multicolumn{4}{c|}{ActivityNet} & \multicolumn{4}{c|}{DiDeMo} & \multicolumn{4}{c}{VATEX}\\
    & & R@1 $\uparrow$ & R@5 $\uparrow$ & R@10 $\uparrow$ & MnR $\downarrow$ & R@1 $\uparrow$ & R@5 $\uparrow$ & R@10 $\uparrow$ & MnR $\downarrow$ & R@1 $\uparrow$ & R@5 $\uparrow$ & R@10 $\uparrow$ & MnR $\downarrow$\\
    \midrule
    & CLIP4Clip~\cite{luo2022clip4clip}                    
    & 40.5 & 72.4 & 98.1 & 7.5 & 43.4 & 70.2 & 80.6 & 17.5 & 55.9 & 89.2 & 95.0 & 3.9 \\
    \midrule
    \multirow{5}{*}{\rotatebox{90}{Adapter}} 
    & Bias~\cite{cai2020tinytl} 
    & 31.3 & 60.3 & 74.2 & 13.4 & 36.5 & 63.4 & 75.2 & 24.8 & 52.2 & 83.1 & 91.3 & 5.2\\
    & Adapter\textsuperscript{ATTN}~\cite{he2022towards} 
    & 31.6 & 60.5 & 74.4 & 13.1 & 36.4 & 62.8 & 73.9 & 23.5 & 52.6 & 83.4 & 91.8 & 5.0\\
    & Adapter\textsuperscript{FFN}~\cite{chen2022adaptformer} 
    & 31.8 & 61.0 & 75.0 & 12.8 & 36.3 & 63.4 & 75.4 & 22.9 & 52.3 & 83.3 & 91.5 & 5.2\\
    & Visual-Text Adapter~\cite{houlsby2019parameter} 
    & 33.5 & 64.8 & 77.5 & 10.9 & - & - & - & - & 53.1 & 85.0 & 92.3 & 4.9\\
    & Video-Text Adapter~\cite{pan2022st} 
    & 36.4 & 66.1 & 79.6 & 10.0 & - & - & - & - & 53.5 & 85.0 & 92.4 & 4.7 \\
    \midrule  
    \multirow{6}{*}{\rotatebox{90}{Prompt}}       
    & VoP\textsuperscript{F+C}~\cite{huang2023vop} 
    & 36.1 & 65.5 & 78.5 & 10.9 & 45.3 & \textcolor{blue}{\textbf{72.3}} & 80.4 & 13.8 & 54.2 & 85.2 & \textcolor{red}{\textbf{93.6}} & 4.7\\          
    & DGL-Linear~\cite{yang2024dgl} 
    & 38.3 & 68.4 & 79.2 & 10.3 & 44.2 & 70.6 & 80.2 & 15.8 & \textcolor{blue}{\textbf{56.2}} & \textcolor{blue}{\textbf{87.1}} & 93.5 & \textcolor{blue}{\textbf{4.1}}\\
    & DGL-Transformer~\cite{yang2024dgl}         
    & 40.1 & 69.5 & 80.9 & 9.1 & 45.6 & 71.7 & 81.1 & 14.6 & 54.3 & 85.5 & 92.3 & 4.9\\
    & UniPT~\cite{diao2024unipt}
    & 34.6 & 65.7 & 75.2 & 15.5 & 40.1 & 64.2 & 74.7 & 18.7 & 53.4 & 85.2 & 92.3 & 5.0\\
    & MPT-Transformer~\cite{zhang2024mpt}        
    & \textcolor{blue}{\textbf{41.4}} & \textcolor{blue}{\textbf{70.9}} & \textcolor{blue}{\textbf{82.9}} & \textcolor{blue}{\textbf{7.8}} & \textcolor{blue}{\textbf{46.4}} & 72.2 & \textcolor{blue}{\textbf{81.4}} & \textcolor{blue}{\textbf{13.4}} & - & - & - & -\\
    & \textbf{STOP (Ours)} 
    & \textcolor{red}{\textbf{43.1}} & \textcolor{red}{\textbf{71.4}} & \textcolor{red}{\textbf{83.7}} & \textcolor{red}{\textbf{6.9}} & \textcolor{red}{\textbf{47.5}} & \textcolor{red}{\textbf{73.5}} & \textcolor{red}{\textbf{82.0}} & \textcolor{red}{\textbf{12.5}} & \textcolor{red}{\textbf{57.5}} & \textcolor{red}{\textbf{88.4}} & \textcolor{blue}{\textbf{93.2}} & \textcolor{red}{\textbf{4.0}} \\
    \bottomrule
  \end{tabularx}
  \vspace{-7pt}
  \begin{flushleft}
        \footnotesize{- The best and second best results are marked in \textcolor{red}{\textbf{RED}} and \textcolor{blue}{\textbf{BLUE}}, respectively. }
    \end{flushleft}
    \vspace{-11pt}
  \caption{Comparison with state-of-the-art on the ActivityNet, DiDeMo, and VATEX. Briefly, we only report text-to-video retrieval (t2v) results. For a fair comparison, all methods use CLIP-ViT-B/32~\cite{radford2021learning} as the backbone.}
  \label{tab:activity}
  
\end{table*}

\subsection{Comparison Methods}
We compare our STOP with both adapter-based and prompt-based parameter-efficient finetuning methods. We also report the fully tuning results as a baseline, i.e., CLIP4Clip~\cite{luo2022clip4clip}. For adapter-based methods, we report the results of  Bias~\cite{cai2020tinytl}, Adapter\textsuperscript{ATTN}~\cite{he2022towards}, Adapter\textsuperscript{FFN}~\cite{chen2022adaptformer}, Visual-Text Adapter~\cite{houlsby2019parameter} and Video-Text Adapter~\cite{pan2022st}. For prompt-based methods, we report the results of Efficient Prompt~\cite{ju2022prompting}, VPT~\cite{jia2022visual}, UPT~\cite{zang2022unified}, VoP\textsuperscript{F+C}~\cite{huang2023vop}, DGL-Linear~\cite{yang2024dgl}, DGL-Transformer~\cite{yang2024dgl}, UniPT~\cite{diao2024unipt}, MPT-Linear~\cite{zhang2024mpt} and MPT-Transformer~\cite{zhang2024mpt}.

\subsection{Implementation Details}
Following \cite{yang2024dgl,zhang2024mpt}, the video and text encoders are initialized with pre-trained CLIP (ViT-B/32), with pre-trained weights kept frozen during training. All video frames are resized to 224×224 and divided into 49 non-overlapping patches. For video action recognition, we use a handcrafted prompt template ``a video of a person doing [CLASS]'' to construct the text input. For the MSR-VTT and VATEX datasets, the maximum length of captions is set to 32, and each video is uniformly sampled to 12 frames. For other datasets, the maximum numbers of sentences and frames are both set to 64. In our method, the default number of intra-frame prompts \(N_s\) is set to 6, the scaling factor of inter-frame prompts \(\eta\) is set to 12, and the weight hyper-parameters \(\alpha\) and \(\beta\) are set to 0.4 and 4, respectively. During training, the model is trained with the AdamW optimizer, with a cosine decay scheduler. 

\subsection{Comparison with State-of-the-arts}
\subsubsection{Action Recognition}
We first validate the effectiveness of our method on the video action recognition task, conducting experiments on HMDB51, UCF101, and SS-V2. As shown in Table~\ref{tab:action}, compared to other state-of-the-art parameter-efficient fine-tuning and video prompting methods, our STOP achieves improvements of \textbf{2.2\%}, \textbf{1.7\%}, and \textbf{3.1\%} on the three datasets, respectively. This is because our STOP leverages integrated spatial-temporal dynamic prompting, which guides the pre-trained model to focus on discriminative regions with temporal dynamics. This approach enhances the pre-trained CLIP model's ability to capture fine-grained temporal information, leading to more accurate action recognition results.

\subsubsection{Video-Text Retrieval}
Next, to further validate the effectiveness of our method, we conducted experiments on the video-text retrieval task. As shown in Table~\ref{tab:msrvtt}, on the MSR-VTT dataset, compared to the second-best method MPT-Transformer, our method achieves a reduction of \textbf{2.08 MB} in parameters while improving the mean rank (MnR) by \textbf{0.4\%} for text-to-video and \textbf{0.5\%} for video-to-text, with R@1 increasing by \textbf{1.4\%} and \textbf{1.1\%}, respectively. Additionally, our method consistently shows improvements in the R@5 and R@10 metrics. This is because our proposed intra-frame spatial prompts and inter-frame prompts complement each other, highlighting the locations of discriminative regions and keyframes in both spatial and temporal dimensions, which facilitates the model in accurately extracting video representations.

Furthermore, as shown in Table~\ref{tab:activity}, we also conducted experiments on datasets ActivityNet, DiDeMo, and VATEX. Compared to the MPT-Transformer, our STOP method on dataset ActivityNet improves R@1, R@5, and R@10 by \textbf{1.7\%}, \textbf{1.5\%}, and \textbf{0.8\%}, respectively, while decreasing the MnR by \textbf{0.9\%}. On the other two datasets, DiDeMo and VATEX, our method also achieves consistent performance improvements. This is because the positions of our intra-frame prompts and the inter-frame prompts are dynamically adjusted, allowing our method to generalize well across different datasets.

\begin{table}
    \centering

    \begin{tabular}{cc|ccc}
    \toprule
    Intra & Inter & HMDB51 & UCF101 & SS-V2\\    
    \midrule
        - & -                                    & 50.2 & 70.2 & 10.2\\
        \ding{51} & -                            & 67.1 & 83.1 & 16.5\\
        - & \ding{51}                            & 69.4 & 87.4 &15.9\\
        \ding{51} & \ding{51}                    & \textcolor{red}{\textbf{72.0}} & \textcolor{red}{\textbf{95.3}} & \textcolor{red}{\textbf{21.4}}\\
    \bottomrule
    \end{tabular}
    \caption{\label{tab:components}
    Ablation study about the influence of components in STOP. ``-'' and ``\ding{51}'' represent without or with this component. ``Intra'' represents the intra-frame spatial prompting and ``Inter'' represents the inter-frame temporal prompting.}
\end{table}

\subsection{Ablation}
\subsubsection{Influence of Different Components}
To verify the effectiveness of the intra-frame spatial prompting and inter-frame temporal prompting, we conducted ablation experiments on three datasets: HMDB51, UCF101, and SS-V2, as shown in Table~\ref{tab:components}. As demonstrated, when neither component is used, STOP degrades to a pre-trained CLIP model. When the intra-frame spatial prompting is used alone, the model's performance improves by 16.9\% on HMDB51. This is because the intra-frame spatial prompts highlight key regions with significant dynamic changes for video understanding, facilitating the pre-trained CLIP model to accurately capture fine-grained temporal information. When the inter-frame temporal prompting is used alone, the model's performance improves by 19.2\%. This is due to the inter-frame prompt capturing temporal dynamics between frames and guiding the pre-trained model to focus on keyframes in the video. When both components are used together, the model's performance improves by an additional 2.6\%, as they complement each other by guiding the pre-trained model to focus on discriminative regions in both the spatial and temporal dimensions, enhancing the model’s ability to understand temporal dependencies.

\begin{figure}[tbp]
\centering
\includegraphics[width=1\linewidth]{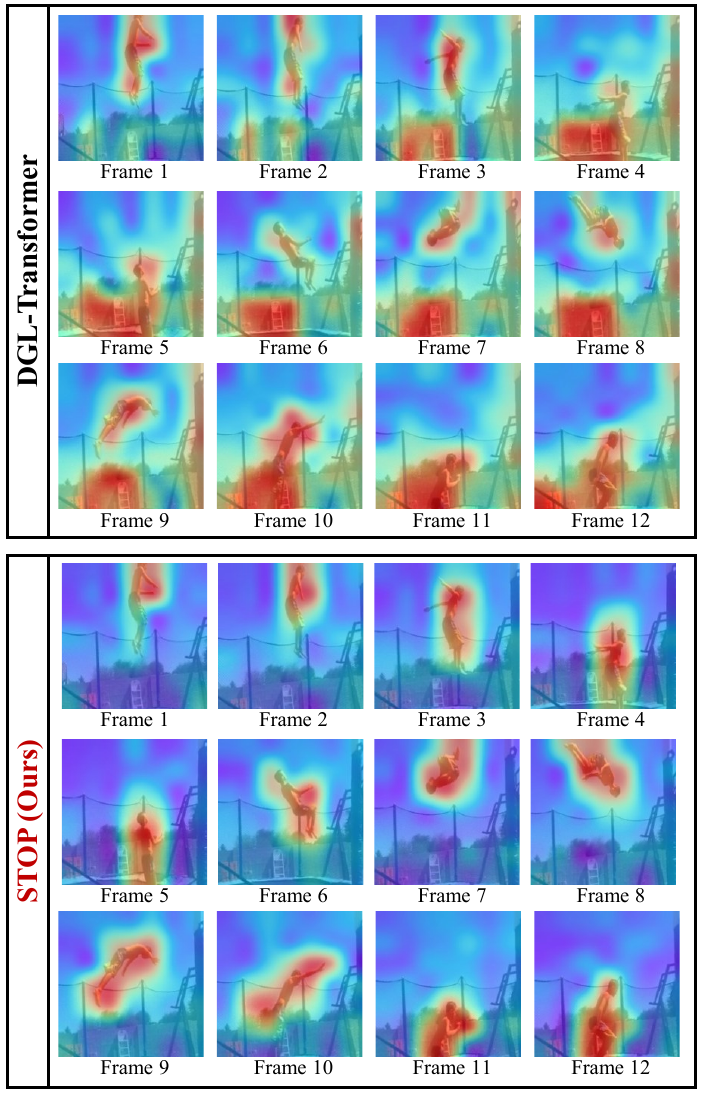}
\caption{\label{fig:attention} Attention map visualization results of existing method DGL and STOP (Ours). Visualization of more cases is included in the supplementary material.}
\end{figure}

\subsubsection{The Visualization Results of Attention Map}
To further explore the impact of our intra-frame spatial prompting and inter-frame temporal prompting, we visualize the attention maps of each video frame. As shown in Figure~\ref{fig:attention}, existing video prompting methods like DGL-Transformer use the same static prompt for all videos. This causes the pre-trained CLIP model to focus on static objects in the video, such as chairs, basketball hoops, and poles. As a result, the model struggles to accurately understand the actions performed by people in the video. 
In contrast, our intra-frame spatial prompting and inter-frame temporal prompting highlight the key regions with dynamic changes in the video. This enables the pre-trained model to focus on the person and his actions in the video, leading to a more accurate understanding.

\begin{figure}
\centering
\includegraphics[width=1\linewidth]{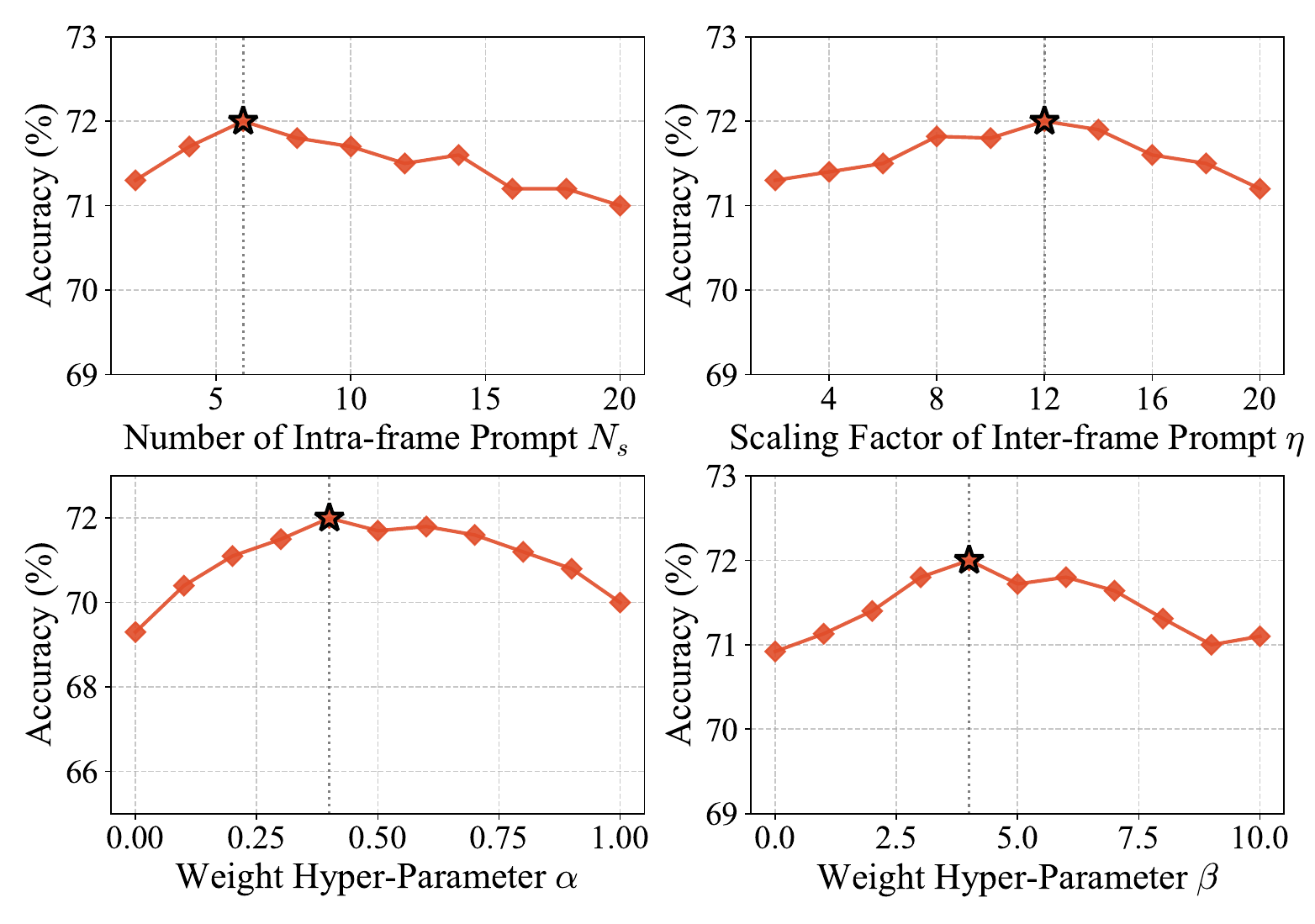}
\caption{\label{fig:hyper}
 Influence of hyper-parameters of STOP in HDMB51.}
\end{figure}

\subsubsection{Influence of Hyper-parameters}
We conduct ablation experiments to investigate the impact of several hyper-parameters in our method, including the intra-frame prompt number \( N_s \), the scaling factor of inter-frame prompts \( \eta \), and the weight hyper-parameters \( \alpha \) and \( \beta \). The results, as illustrated in Figure~\ref{fig:hyper}, indicate that the model performs best when \( N \) is set to 6. This is because, when \( N_s \) is too small, the intra-frame spatial prompts fail to cover the discriminative regions, while if \( N_s \) is too large, it leads to an overemphasis on the background. For the scaling factor \( \eta \), when it is too small, the number of inter-frame temporal prompt tokens is insufficient, limiting the model's learning capacity. On the other hand, when \( \eta \) is too large, the excessive number of inter-frame temporal prompts interferes with feature extraction. The optimal performance is achieved when \( \eta \) is set to 12. 

The weight hyper-parameter \( \alpha \) controls the importance of the intra-frame attention map and the temporal dynamics in the inter-frame prompt when computing the discriminative regions. When \( \alpha \) is set to 0.4, it balances both aspects, leading to more accurate identification of discriminative regions and achieving the best performance. Similarly, the weight hyper-parameter \( \beta \) affects the importance of discriminative regions when computing the temporal dynamics across frames. When \( \beta \) is set to 4, it strikes a balance and results in optimal model performance.

\section{Conclusion}
\label{sec:conclusion}
In this paper, we propose a novel integrated spatial-temporal dynamic prompting (STOP) approach. Specifically, we design intra-frame spatial prompts with dynamically changing positions, as well as inter-frame temporal prompts with dynamically changing quantities based on the temporal variations of the video. This approach guides the pre-trained vision-language model to focus on discriminative spatiotemporal regions, enhancing the model's ability to accurately understand temporal dependencies within the video. The effectiveness of our proposed STOP has been validated on multiple large-scale benchmarks for video action recognition and video-text retrieval.

\section*{Acknowledgments}
This work was supported by the grants from the National Natural Science Foundation of China (62376011, 61925201, 62132001, 62432001) and Beijing Natural Science Foundation (L247006).

{
    \small
    \bibliographystyle{ieeenat_fullname}
    \bibliography{main}
}

\clearpage
\setcounter{page}{1}
\maketitlesupplementary

\begin{figure}[t]
\begin{center}
\includegraphics[width=1\linewidth]{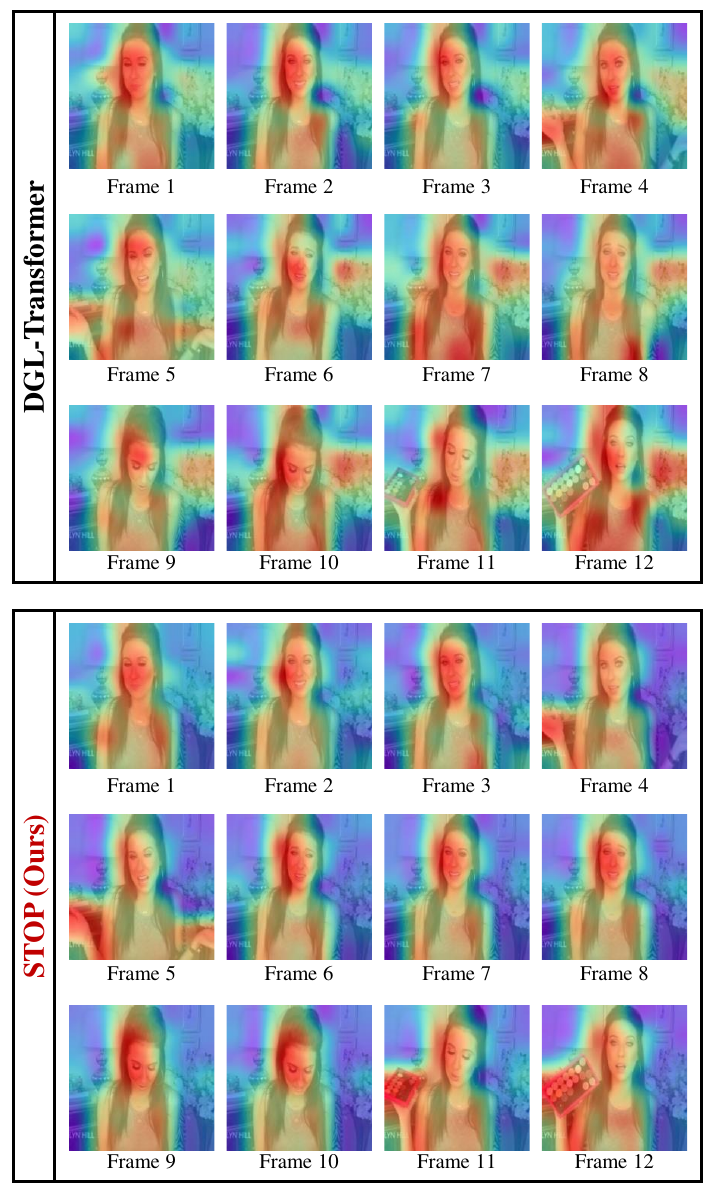}  
\end{center}
\caption{\label{fig:supp-attention-2} 
More Attention map visualization results of existing method DGL-Transformer~\cite{yang2024dgl} and STOP (Ours).
}
\vspace{10pt}
\end{figure}


\begin{figure}[t]
\begin{center}
\includegraphics[width=1\linewidth]{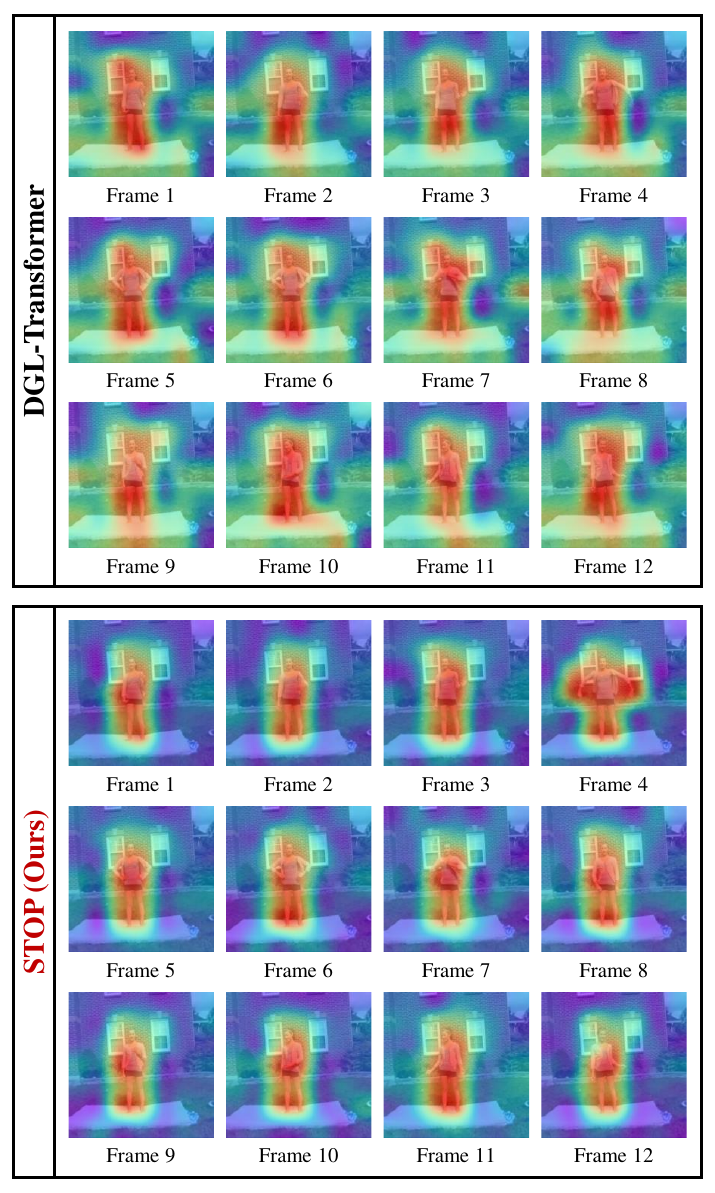}  
\end{center}
\caption{\label{fig:supp-attention-1} 
More Attention map visualization results of existing method DGL-Transformer~\cite{yang2024dgl} and STOP (Ours).
\vspace{10pt}
}
\end{figure}

\begin{figure*}[htbp]
\begin{center}
\includegraphics[width=0.9\linewidth]{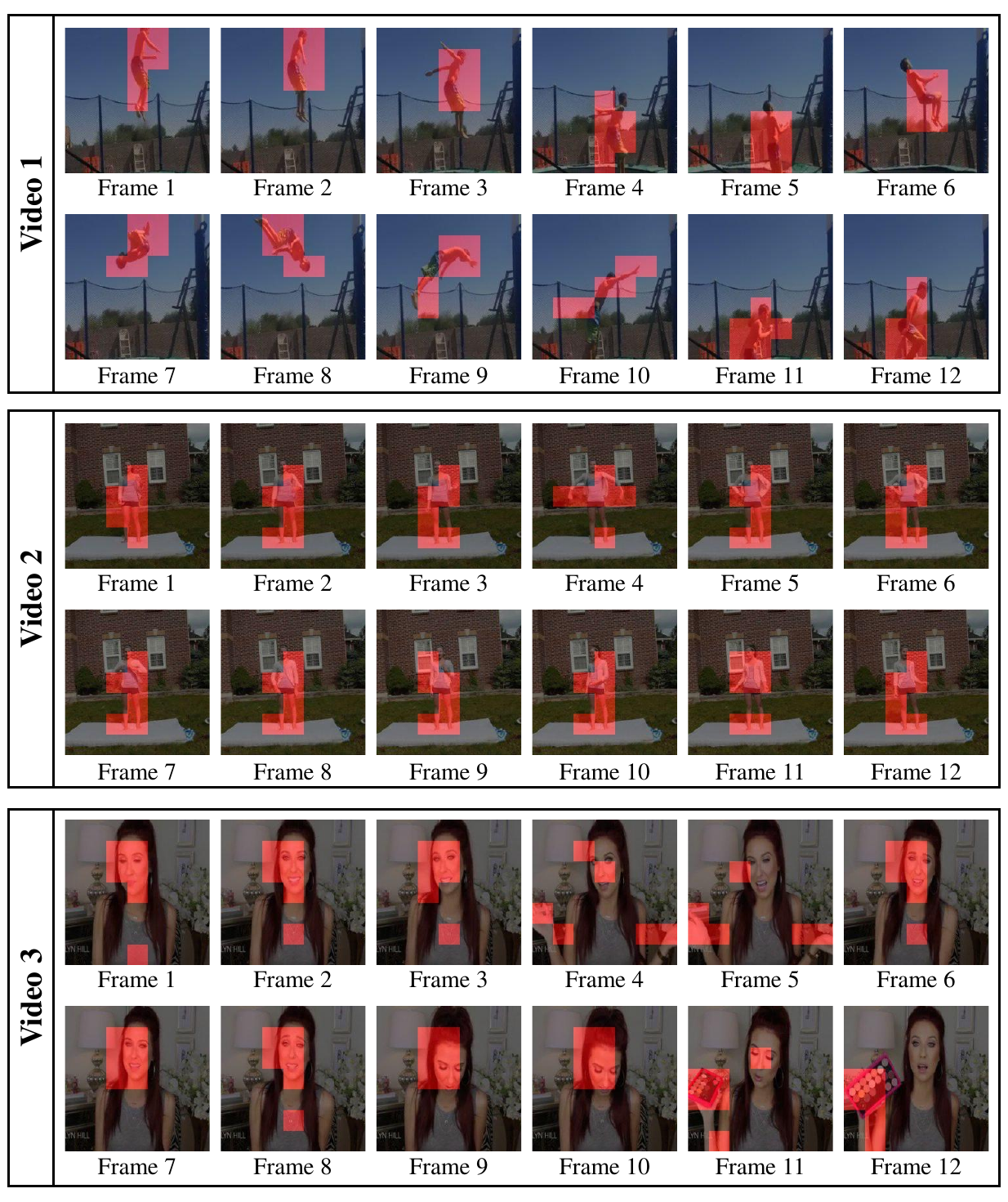}  
\end{center}
\caption{\label{fig:supp-prompt} 
The visualization results of the positions where the intra-frame spatial prompt is added.
}
\end{figure*}

\section{Attention Map Visualization of More Cases}
To further explore the impact of our intra-frame spatial prompting and inter-frame temporal prompting, we visualized the attention maps for more video cases. As shown in Figure~\ref{fig:supp-attention-1} and Figure~\ref{fig:supp-attention-2}, existing video prompting methods (e.g., DGL-Transformer) use the same static prompt for all videos. This causes the pre-trained CLIP model to focus on static objects and backgrounds in the video, making it challenging to accurately understand the actions of people in the video.  

In contrast, our intra-frame spatial prompting and inter-frame temporal prompting highlight the key regions with dynamic changes in the video, enabling the pre-trained model to focus on the people and their actions. This leads to a more accurate understanding and shows a similar trend to the visualized results presented in the main text.

\section{Visualization of Intra-Frame Spatial Prompt}
To verify the effectiveness of our intra-frame spatial prompting, we visualized the positions where it is added. As shown in Figure~\ref{fig:supp-prompt}, the red patches indicate the locations where the intra-frame spatial prompt is applied. It can be observed that our method comprehensively considers intra-frame attention weights and temporal variations, enabling accurate localization of discriminative regions in the video. This facilitates the pre-trained vision-language model to focus accurately on these discriminative regions, enhancing the model's ability to extract temporal information.



\end{document}